# Article citation study: Context enhanced citation sentiment detection

Vishal Vyas[a], Kumar Ravi[b,*], Vadlamani Ravi[c], V.Uma[a], Srirangaraj Setlur[d], Venu Govindaraju[d]

[a]Department of Computer Science, School of Engineering and Technology, Pondicherry University, Pondicherry 605014, India
[b]HCL Technologies Ltd., Noida, India, 201301
[c]Centre of Excellence in Analytics, IDRBT, Hyderabad, India, 500057
[d]Center for Unified Biometrics and Sensors 113 Davis Hall, University at Buffalo, Amherst, NY, 14260-2500

## Abstract

Citation sentimet analysis is one of the little studied tasks for scientometric analysis. For citation analysis, we developed eight datasets comprising citation sentences, which are manually annotated by us into three sentiment polarities viz. positive, negative, and neutral. Among eight datasets, three were developed by considering the whole context of citations. Furthermore, we proposed an ensembled feature engineering method comprising word embeddings obtained for texts, parts-of-speech tags, and dependency relationships together. Ensembled features were considered as input to deep learning based approaches for citation sentiment classification, which is in turn compared with Bag-of-Words approach. Experimental results demonstrate that deep learning is useful for higher number of samples, whereas support vector machine is the winner for smaller number of samples. Moreover, context-based samples are proved to be more effective than context-less samples for citation sentiment analysis.

## 1 Introduction

Tremendous studies are carried out for scientometric analysis. However, the majority of indexing services consider quantitative citation to determine the usefulness of a research article. A few studies are carried out to determine qualitative citation. In this regard, citation sentiment analysis (CSA) is one of the little explored methods to understand the quality of citation (Athar, 2011). CSA is a process to identify the author's criticism and appreciation towards a cited paper. The goal of CSA is to identify the semantic orientation of each cited work in a research paper. The emotion/critics behind citation indicates the usefulness of the publication.

The sentiment in addition to the citation score would help the research community to search the relevant article conveniently. If a publication has high negative sentiment score, it implies that there are some serious issues with the research. Other researchers are citing the stated paper for its flaws. An availability of such sentiment score along with the number of citations would make the relevant paper accessible more effectively. Considering highly cited research article might not be a good idea always. There might be a possibility of having a high negative sentiment associated with the article. The high negative sentiment clearly reveals the non-reproducibility or ill-facts about the research.

Our assumption stems from the observation that, besides using citations as a background for their current work, authors often take positive/negative instances towards the cited work. For example, an author may approve previous work and cite it as supporting evidence for his/her own statements or points, or cite it as a negative example to be criticized in his/her article, as shown in Table 1. Table shows three citation contexts extracted from article (Maks & Vossen, 2012), which are manually annotated with three polarities viz. *positive*, *negative* and *neutral* towards a cited work. The first column presents the citing paper, the second column represents citation context which we obtained after going through the paper, the third column presents the cited work i.e. target article, and the last column presents the manually annotated polarity.

**Table 1: Examples of positive, negative and neutral citations in Elsevier paper id - S0167739X13001349**

| Citing paper ID | Citation window including context | Cited Work | Polarity |
|---|---|---|---|
| S01677 39X13001 349 | Aggarwal and Vitter proposed the Disk Access Machine (DAM) model [22] which counts the number of memory transfers from slow to fast memory instead of simply counting the number of memory accesses by the program. Therefore, it better captures the fact that modern machines have memory hierarchies and exploiting spatial and temporal locality on these machines can lead to better performance. There are also a number of other models that consider the memory access costs of sequential algorithms in different ways [23-29]. | [22] | Positive |
| S01677 39X13001 349 | Zhang and OwensÃ, [15] present a quantitative performance model that characterizes an application performance as being primarily bounded by one of three potential limits: instruction pipeline, shared memory accesses, and global memory accesses. More recently, Sim etÃ. al.Ã, [48] develop a performance analysis framework that consists of an analytical model and profiling tools. The framework does a good job in performance diagnostics on case studies of real codes. Kim et al. [49] also design a tool to estimate GPU memory performance by collecting performance-critical parameters. Parakh et al. [50] present a model to | [15], [48], [49] | Negative for all three citations |
| | estimate both computation time by precisely counting instructions and memory access time by a method to generate address traces. All of these efforts are mainly focused on the practical calibrated performance models. No attempts have been made to develop an asymptotic theoretical model applicable to a wide range of highly-threaded machines. | | |
| S016773 9X13001 349 | To motivate this enterprise and to understand the importance of high thread counts on highly-threaded, many-core machines, let us consider a simple application that performs Bloom filter set membership tests on an input stream of bio sequence data on GPUs [3]. | [30] | Neutral |

CSA mainly determines the appreciation and criticism of the existing study. (Athar, 2011) is the first to analyze sentiment polarity of a sentence (of an article), in which a citation is occurring. He devised three rules to determine an appreciation and a criticism of the existing study (Athar, 2014b). In the annotation scheme, the citation sentences which criticized the results, approach or performance of the cited work were marked with negative polarity. The citations that confirm, support or use the existing cited work were considered under positive polarity. The citations that were neither negative nor affirmative, were considered to be marked as neutral.

CSA can have various possible applications. Some of them can be listed as follows: (i) the number of positive and negative mentions can be included in the calculation of citation metric like impact factor, cite score, etc. (ii) CSA can also be useful in citation bias analysis (Athar, 2014b). (iii) Automatic CSA can be used to figure out a list of studies, which are not reproducible. (iv) Article recommendation system can be leveraged using citation sentiment too (Ravi et al., 2018). In this study, we extended the work carried out by (Ravi et al., 2018). They performed sentiment classification for single citation sentences present in two datasets. The first dataset was taken from (Athar, 2011) and the second citation dataset was developed from the articles collected from http://www.sciencedirect.com by the first author of (Ravi et al., 2018). (Ravi et al., 2018) raised some of the major issues in CSA are high imbalance data, granularity issue, the vagueness of valence, multiple citations, and use of indirect narration. In order to tackle granularity and vagueness of valence issues, we collected the real context of each citation by going through research articles thoroughly. We developed two context-enhanced datasets in order to resolve the above-mentioned issues and employed traditional machine learning as well as deep learning techniques to predict the citation sentiment. We performed sentiment classification of article citation sentences into three categories viz. positive, negative, and neutral. Due to the scarcity of negative citation sentences, the corpus suffers from huge imbalance issue.

## 1.1 Objectives

In this study, we worked towards following objectives:

- We wanted to understand the effects of context on citation sentiment classification compared to single citation sentence.
- We have observed that considering embedding of dependency relations along with word embedding is helpful for sentiment classification (Ravi et al., 2018). We considered to exploit this novel ensembled features for more citation datasets.
- We would like to compare the performance of recently proposed region-embedding based deep learning methods (Johnson & Zhang, 2014), (Johnson & Zhang, 2016) with traditional machine learning methods for citation sentiment classification.

To fulfill these objectives, we developed two context-based corpora by going thorough 70 articles from www.sciencedirect.com to create 2164 samples and another 70 articles from http://cl.awaisathar.com/citation-sentiment-corpus to create 1874 samples. Unlike (Athar, 2011), we considered the real context for annotation. To the best of our knowledge, no citation corpus has been developed by considering the real context of the citation text. We performed context based annotation with three sentiment polarities viz. positive, negative and neutral.

The rest of this paper is organized as follows. Section 2 reviews the related work. We present the ensembled feature selection method for classification using deep learning and machine learning approaches in Section 3. We then describe the data and experimental setup with hyperparameter tuning details in Section 4. Section 5 presents results and discussion. Finally, section 6 concludes the paper and suggests directions for future work.

## 2 Related work

Quality assessment of a large number of scientific papers is a cumbersome task. Over the years different practices to study citation patterns of bibliometric measures have increased the interest of many authors. But, it is rather quantitative as it depends on the citation count. Qualitative aspect can be measured by considering the citation context which in turn will help in improving bibliometric measures. (White, 2004) surveyed the study of citation in different research directions. (Athar, 2014b) introduced three rules to decide criticism and appreciation of a citation in the existing study. These rules guided the annotation schemes to label citation sentences into positive, negative and neutral polarities. They performed citation context detection, citation polarity determination and citation purpose detection. In the experiment, they considered manually annotated 3500 citation sentences. They employed conditional random field for citation context detection and yielded F-score of 89.5%. For the rest two tasks, suppport vector machine (SVM) yielded macro-F1 of 58% and 62% for purpose detection and polarity detection respectively.

(Radev, Muthukrishnan, Qazvinian, & Abu-Jbara, 2013) applied supervised sequence labelling technique to find citation context and binary classification. Due to different reference styles, the authors used a regular expression to get obtain 4 sentences based citation context. SVM yielded macro-F1 score of 71%. (Sula & Miller, 2014) scrutinized the constraints of citation context in the humanistic discourse. They developed a tool to extract and classify the citation context in the journals published in the humanities domain. They claimed that the statistics of citation figures in the performed experiment was consistent compared to the previous researches in the same domain. (Parthasarathy & Tomar, 2014) utilized sentence parser to select citation sentences from arbitrary papers. Based on availability of adjectives, they classified citation sentences into three categories viz. positive, negative, and neutral. The experiment was conducted on a smaller dataset.

The author (Munkhdalai et al., 2016) introduced a compositional attention network for citation classification and sentiment analysis. They prepared a dataset comprising 5,000 citation sentences randomly collected from 2,500 PubMed central articles. They considered one preceding and one succeeding sentence of the citation sentence. They achieved macro-F1 of 75.67% and 78.1% for citation function and citation sentiment classification respectively. In these studies, the prepared datasets were based on fixed number of sentences, which did not capture the complete context of the cited work. The author (Lauscher, Glavaš, Ponzetto, & Eckert, 2017) employed word vector CNN on a dataset containing 3,271 citation context instances. Each citation context consists of four sentences viz. the sentence containing the citing work, one preceding sentence and two following sentences. They performed citation polarity and citation purpose detection and reported macro-F1 of 78.8% and 74.3% for respectively. The authors (Karyotis, Doctor, Iqbal, James, & Chang, 2018), (Munkhdalai, Lalor, & Yu, 2016) and

(Ravi et al., 2018) employed deep learning models. In these studies, Convolutional Neural Network (CNN) and Long Short-Term Memory (LSTM) were used to perform citation sentiment classification. The author (Ravi et al., 2018) utilized various deep learning approaches for citation sentiment classification. They experimented with 8,925 and 6,567 single citation sentences collected from Elsevier and ACL scientific articles and reported macro-F1 of 62.99% and 54.49% respectively.

(Bu, Waltman, & Huang, 2019) introduced a multidimensional model for categorizing citation impact of a publication into two groups. The first group includes publications having a deep impact toward a specific field. The second group comprises publications having impact beyond one specific research field. Furthermore, they also categorized publications having strongly dependent on previous work and publications making a fundamental contribution. They considered 36.2 million publications appeared during 1980 to 2017 to develop the dataset, which covered 699.3 million citation relations between these publications. (Wang, Leng, Ren, Zeng, & Chen, 2019) introduced Conditional Random Field (CRF) based model to annotate the logical relation between syntactic structure and vocabulary in linguistic patterns. They presented the role of linguistic patterns in classifying the citation sentiments. They observed that linguistic patterns can improve sentiment classification of citation context. (Ikram & Afzal, 2019) determined aspect level sentiment to reveal the hidden patterns in publications. The experiments were performed on two different datasets. The first dataset consists of 8736 citation sentences from ACL anthology Network. The second dataset were containing 4182 citation sentences from clinical trial papers. They employed various machine learning for classification and reported an F-measure of 75.5%. (Zhao & Strotmann, 2020) introduced location filter citation counting method. Their counting method filters out citation from introduction and background sections of the article. It then weighs remaining citations based on their in-text frequency. They experimented with bibliometrics extracted from PubMed central. They observed that removing introductory and background citations do not make much difference.

Based on the reviewed literature, we can observe that majority of the studies are either carried out on fixed window of sentences or single sentence. However, the real context would be captured by considering all the sentences related to citation. Further, context based sentiment analysis would yield better accuracy for citation sentiment classification. Hence, we considered to work on context based citation sentences.

# 3 Proposed Approach

This work is an extended work of (Ravi et al., 2018), wherein, they proposed a novel feature generation method to classify citation sentiment corpus using deep learning approaches. They employed different architectures of the CNN and LSTM for citation sentiment classification. In addition to the collection of words from the citation corpus, they considered dependency relations of citation sentence sentiment as an input to deep learning models. The performance of these models are compared with the performance of Bag-of-Words (BoW) models. Based on extensive experiments, authors indicated that a bigger citation window may lead to the improved performance of the deep learning approaches (Ravi et al., 2018). Therefore, in this paper, we considered to experiment with context enhanced citation samples. For Bag-of-Words (BoW) models, we generated Term Frequency-Inverse Document Frequency (TF-IDF) based Document Term Matrix (DTM) to classify citation sentences into three classes. The deep learning and BoW models based approaches are illustrated in Figure 1 and Figure 2 respectively.

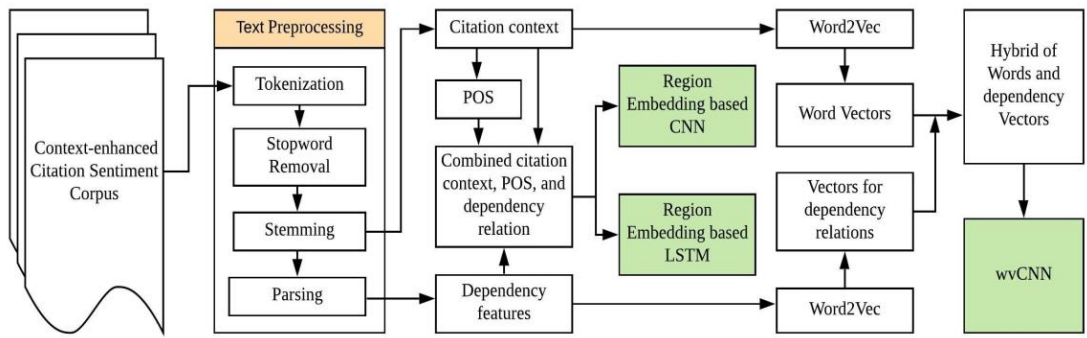

Figure 1: The deep learning based approach

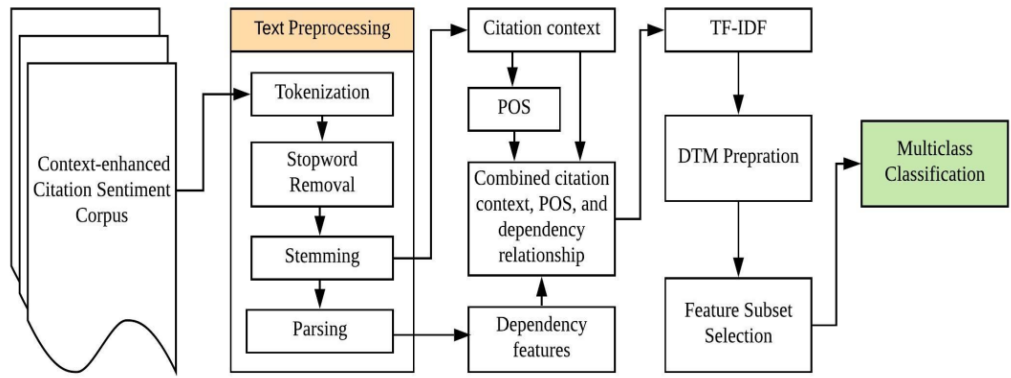

Figure 2: The Bag-of-Words based approach

## 3.1 Text preprocessing

We performed text preprocessing steps namely tokenization, stop word removal, and stemming for BoW models. Firstly, all words were converted into tokens and then stop words like *'is'*, *'am'*, *'are'*, *'of'*, *'the'* etc. were removed from the corpus. Words such as *'considered'*, *'considering'* were converted to its root form *'consider'*, and this process is known as stemming. For parsing, we employed Stanford parser (Toutanova, Klein, Manning, & Singer, 2003) which generates the dependency relations. Same prepocessing steps were applied to all the datasets. We have utilized the snowball stemmer, stop word removal (English) of NLTK (Loper & Bird, 2002).

## 3.2 Feature Extraction

For BoW models, we considered the ensemble features comprising content, syntactic and semantic features. For content based features, we considered n-grams of the text. For syntactic features, Part-of-Speech tags (POS) were considered. For semantic features, dependency relations of the sentences were considered. Each dependency relation contains three components namely dependency relation name, governor and dependent to form a single word. For example, *nusbj(good-1, cite-2)* was converted as *nsubj_good_cite*. We concatenated POS tags and dependency relations of the text at the end of the sample citation text. After combining, content, syntactic, and semantic features, a sample example looks like below:

```
"others perform division implicitly without discussing
```

```
performance eg cutting et al root_root_-lsb- nsubj_perform_others
ccomp_-lsb-_perform    dobj_perform_division    advmod_perform_
implicitly    mark_discussing_without    advcl_perform_discussing
compound_eg_performance    dobj_discussing_eg    nmod_poss_-rsb-
_cutting   cc_cutting_et   conj_et_cutting_al   nmod_poss_-rsb-_al
case_cutting_ńsubj_-lsb-_-rsb- others_NNS perform_VBP division_NN
implicitly_RB   without_IN   discussing_VBG   performance_NN   eg_NN
cutting_VBG et_CC al_JJ"
```

DTM was prepared to consider different n-grams. We extracted five different combinations of n-grams namely (i) unigrams, (ii) bi-grams, (iii) tri-grams, (iv) unigrams and bi-grams together, and (v) unigrams, bi-grams and tri-grams together. In order to reduce the number of features, we applied chi-square ($\chi^2$) and information gain based feature subset selection.

For deep learning models, we extracted word vectors for all *n*-grams using pre-trained word embedding viz. glove.840B.300d[1]. The vectors for words and dependency relations, which were not available in GloVe pre-trained word vector space, were initialized randomly as given in (Kim, 2014). In order to train CNN, we appended the word vector of each of the dependency relations and POS tags at the end of all unigrams. Each row of the matrix of the input layer is a vector of a word or a dependency relation. In order to train region embedding based CNN and LSTM, we concatenated the given sentence with the list dependency (grammatical) relations obtained for the sentence. The author (De Marneffe & Manning, 2008) has explained briefly a list of 50 grammatical relations which were used in our work. The concatenated sentence was given as an input to different neural networks.

### 3.3 Deep learning approach

We employed different deep learning architectures for sentiment classification purpose. Employed deep learning architectures include word vector based CNN (wvCNN) (Kim, 2014) one-hot CNN (Oh-CNN) (Johnson & Zhang, 2014) and one-hot bidirectional LSTM with pooling (Oh-biLSTMp) (Johnson & Zhang, 2016). Here, one-hot deep learning based models are also called as region-embedding based models. We employed CNN because it performs feature engineering automatically. Further, final sentiment of the citation window would be very much dependent on all the sentences written about the given citation. In this case, LSTM would be quite useful to consider the context propagating from beginning to end of the context window.

---

[1] https://nlp.stanford.edu/projects/glove/

### 3.3.1 Word vector based convolutional neural network

We experimented with CNN model proposed by (Kim, 2014). CNN contains four layers namely embedding layer, convolutional layer, pooling layer, and fully connected layer. Each word of a sentence is encoded as a vector to be provided as input to CNN. Each row of the input materix corresponds to one word. N-grams determines the filter size and different n-grams are considered to capture opinion features appearing in different window of words. Padding with zero was performed to have same number of rows of the input matrix. In the next step, the embedding layer transforms each word into a low-dimensional dense vector.

(Kim, 2014) proposed four different architectures for low dimensional representation of a word and they are: (i) CNN-rand, where word embeddings were randomly initialized and modified during training, (ii) CNN-static, where pre-trained vectors obtained using word2vec were used. They are kept static throughout the training whereas, other parameters are learned, (iii) CNN-non-static, where pre-trained vectors obtained using *word2vec* are used and also pre-trained vectors are fine-tuned, and (iv) CNN multi-channel, where two or more channels can be used. However, the gradients are back-propagated only through one of the channels. We experimented with all these four architectures with word vectors obtained for each word. For the fourth architecture, the first channel was initialized with vectors of words available in pre-trained vectors on 100 billion words collected from Google News[2] (Mikolov & Dean., 2013). The words, which were not available in Google pre-trained model, were initialized with the uniform random vector in the range of [-0.25, 0.25]. The second channel was initialized with vectors available in *glove.840B.300d* (Pennington, Socher, & Manning, 2014). In the bi-channel architecture, each filter was applied to both the channels, and their outputs were added to obtain a feature, $c_i$.

Out of the four architectures, first and third architectures yielded better results using word vectors. Hence, we considered the first and third architectures for further experiments. Vectors obtained for Words, POS, and dependency triplet were fed as input to these two architectures.

### 3.3.2 One-hot convolution neural network

---
[2] https://code.google.com/p/word2vec/

Oh-CNN is effective for text categorization (Johnson & Zhang, 2014). It is a special case of a general framework which jointly trains a linear model with a non-linear feature generator consisting of text region embedding and pooling layer. Oh-CNN is influenced by the findings of (Kim, 2014). Text categorization through CNN without word2vec layer is not only achievable but produces promising results. Here, the author (Johnson & Zhang, 2014) considered the small regions from the data like *"am so happy"* and converted it into feature vectors. The convolution layer learns an embedding of small regions of text data. Here, there is no need to tune so many hyper-parameters of *word2vec* to obtain the word embedding. The author (Johnson & Zhang, 2014) proposed two region representations viz. BOW-CNN (suitable for topic classification) and SEQ-CNN (suitable for sentiment classification).

In SEQ-CNN region representation, each region is formed by concatenating the vectors of all the words. Here, the word sequence is maintained when text region vectors are converted into low dimensional feature vector space. BOW-CNN works efficiently with large vocabulary size $p \times V$. Region representation with the large $p \times V$ learns few parameters but expressiveness is close to SEQ-CNN. To understand region embedding, let us consider a vocabulary V = {"don't", "hate", "I", "it", "love"}. Let a document, *D*, be "I love it". Then the document vector, is encoded using 1-to-V encoding as $[00100|00001|00010]^T$. Each region is formed by concatenating the vectors of all words. Hence, the dimension of a region of size, will be $p \times V$-dimensional. An architecture of region size, $p$, of 2 and stride of 1 is depicted in Figure 3. A computation unit for convolution is: $\sigma(W.r_x + b)$, here $r_l(x) \in R^q$ is a region vector, $W \in R^{m \times q}$ and $m$ is the number of weight vectors (row of $W$) or neurons. We fixed the number of pooling such that we get a fixed number of features after pooling. Hence, the pooling region size is determined dynamically for each data point (Kalchbrenner, Grefenstette, & Blunsom, 2014). In addition to words of the text and POS tags, we provided the one-hot vector of dependency relations obtained for each sentence.

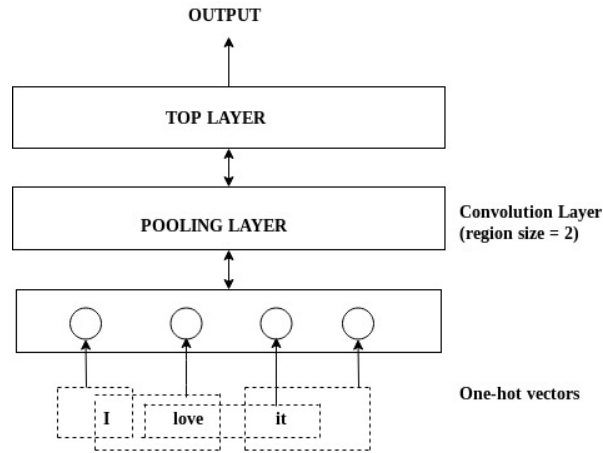

Figure 3: One-hot CNN with a region size 2 and stride 1

### 3.3.3 One-hot bidirectional LSTM with pooling

LSTM is a more sophisticated region embedding method. It can embed text with variable region size. (Johnson & Zhang, 2016) proposed methods to train one-hot LSTM with pooling in supervised settings (Oh-LSTMp), one-hot LSTM with pooling in semi-supervised settings, a hybrid of CNN two-view embeddings, and LSTM two-view embeddings methods. Two-view based models overcome the drawback of one-hot CNN, where the region size is fixed. We employed Oh-biLSTMp, where a fixed number of region embedding is learned from the given document. This in turn is concatenated to obtain document vector. The architecture of the Oh-LSTMp is presented in Figure 4. The presented architecture is a modified form of the architecture proposed in (Zaremba & Sutskever, 2014). They observed that Oh-LSTMp is preferred over word-vector LSTM (wvLSTM) owing to three reasons: (i) wvLSTM underperforms linear predictors, (ii) its training is unstable, and (iii) wvLSTM is time and resource intensive compared to Oh-LSTMp.

In Oh-LSTMp, we input the one-hot vector for each word in the sentence at the first layer. This architecture determines the regions of the text of arbitrary sizes, which in turn is represented as region embedding.

We employed Oh-biLSTMp in our experiment which is depicted in Figure 5. Here, each LSTM cell emits vectors $h_t$ at each time step which are concatenated into a document vector using pooling layer.

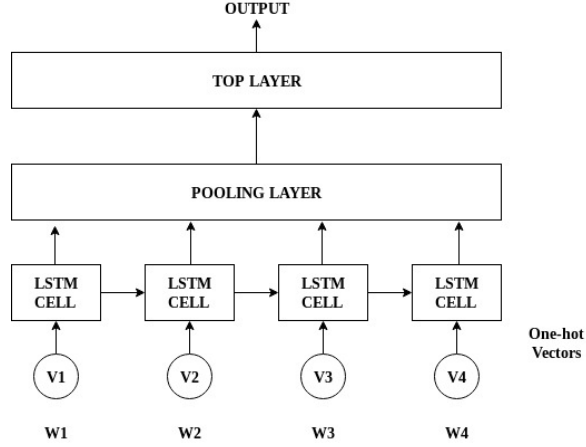

**Figure 4: One-hot LSTM with pooling in supervised setting**

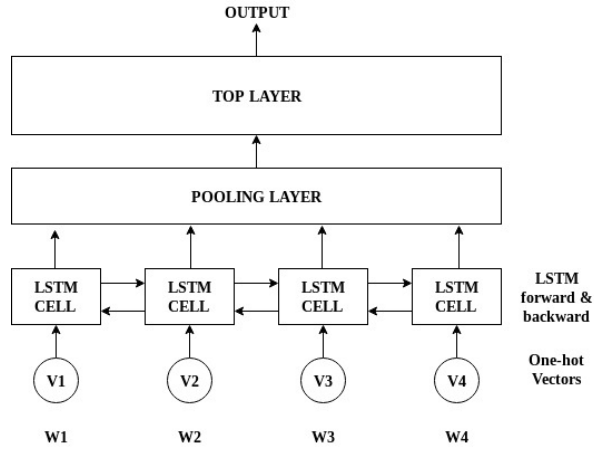

**Figure 5: One-hot bidirectional LSTM with pooling in supervised setting**

In *wvLSTM* setting, each cell needs to propagate the important features till the end of the LSTM layer, whereas Oh-LSTMp captures important phrases or sentences at each cell. LSTM layer helps in faster training of the architecture by chopping each document into multiple non-overlapped segments of fixed length (e.g., 50 or 100). The obtained segments are processed in parallel using mini-batches. Testing is performed without chopping to avoid loss of important content. Finally, input and output gates were not considered in Oh-LSTMp to speed up the process and for better learning. Because, the job of output gate is to avoid entering unnecessary information to $h_t$, which is accomplished using pooling. The formulation of Oh-LSTMp is given as: $f_t = \sigma(W^f x_t + U^f h_{t-1} + b^f)$, $u_t = \sigma(W^u x_t + U^u h_{t-1} + b^u)$, $c_t = u_t + f_t \cdot c_{t-1}$, and $h_t = tanh(c_t)$. Here, '·' is element-wise multiplication operation, $\sigma$ is sigmoid function, $x_t \epsilon R^d$ is the input from the lower layer at time step *t*, and *d* is size of the vocabulary or the dimension of the one-hot vector. For $q$ LSTM units, the dimensionality of the weight vectors are $W^{(\cdot)} \epsilon R^{q \times d}$, $U^{(\cdot)} \epsilon R^{q \times d}$ and $b^{(\cdot)} \epsilon R^q$ for both types of gates viz. *f* and *u*. In addition to words of the text and POS tags, we provided one-hot vector of dependency relations obtained for each sentence.

## 4 Experimental setup

We experimented with context-enhanced citation dataset to analyse the performance of the proposed model. We employed SVM and different deep learning algorithms for classification purpose. Deep learning models include word embedding based CNN models viz. *wvCNN_random*, and *wvCNN_non-static*, and region embedding based viz. Oh-CNN, and Oh-biLSTMp. For word vector based deep learning, word embeddings were obtained using GloVe. For region embedding based models, vectors were generated for regions of the text using deep learning models itself.

### 4.1 Dataset details

We performed multi-class classification considering three classes viz. positive, negative, and neutral. In this study, we experimented with nine datasets developed using citation sentences. Table 2 presents the details of nine datasets, which will be made available at https://github.com/kravi2018/Citation_Sentiment. The first two datasets were taken from the literature and the rest were manually created by us. In the first two datasets, single citation sentences were considered as samples and these datasets is referred to as *context-less* datasets. In the next two datasets, we considered context of the citation sentences to obtain the samples and these datasets will be referred to as *context-full* datasets. To generate a bigger corpora, we combined both *context-less* datasets together to obtain *augmented_context_less*. Similarly, we combined both context-full datasets together to obtain *augmented_context_full*. The rest three datasets contains single citation sentence based samples, which are derived from *context-full datasets*.

Table 2: Distribution of samples of six different datasets

| DATASET | POLARITY | | | TOTAL |
|---|---|---|---|---|
| | POSITIVE | NEGATIVE | NEUTRAL | |
| *elsev_context_less* | 1225 | 181 | 7519 | 8925 |
| *athar_context_less* | 741 | 244 | 5582 | 6567 |
| *elsev_context_full* | 712 | 316 | 1136 | 2164 |
| *athar_context_full* | 571 | 219 | 1084 | 1874 |
| *augmented_context_less* | 1969 | 425 | 13101 | 15492 |
| *augmented_context_full* | 1283 | 532 | 2220 | 4038 |
| *elsev_context_full_derived* | 712 | 316 | 1136 | 2164 |
| *athar_context_full_derived* | 571 | 219 | 1084 | 1874 |
| *augmented_context_full_derived* | 1283 | 532 | 2220 | 4038 |

### 4.1.1 Elsevier dataset without context

The dataset is taken from (Ravi et al., 2018), contains single citation sentences from Elsevier articles. This dataset will be referred to as *elsev_context_less* hereafter. It was developed from 191 open-access articles from www.sciencedirect.com. The author considered only those articles, which were published in journals related to information sciences viz. "Applied soft computing", "Knowledge-based systems", and "Decision support systems". The articles were downloaded as XML, which were preprocessed using ElementTree API[3] to extract citation sentences. The dataset contains a total of 8925 citation sentences out of which positive, negative, and neutral sentences are 1225, 181 and 7519 respectively.

### 4.1.2 Athar dataset without context

The dataset is taken from (Athar, 2011). It can be downloaded from http://cl.awaisathar.com/citation-s entiment-corpus/. The citation sentences were collected from 194 articles from ACL Anthology (Joseph & Radev, 2007), [2]. The author annotated 8736 citations, which were divided into three parts. This dataset will be referred to as *athar_context_less* hereafter. According to (Athar, 2011) and errata mentioned on two websites viz. http://cl.awaisathar.com/citation-sentiment-corpus/ and https://github.com/awaisathar/-CitationSentimentClassifier, the first part contains 736 sentences, which were considered in tuning the parameters of SVM. The second part contains 736 sentences, which were used to develop the science-specific sentiment lexicon. The rest of the citation sentences comprising of 244 negati-9ve, 743 positive, and 6277 objective citations are considered as training and testing sets. Out of the three parts of the data, the first two parts of data were not made available separately. We considered experimenting with 6567 citation sentences, which comprises of 741 positive, 244 negatives, and 5582 neutral citation sentences under the 10-fold cross-validation framework.

### 4.1.3 Elsevier dataset with context

The dataset is manually constructed by us. We considered a bigger citation window from the Elsevier articles. This dataset will be referred to as *elsev_context_full* hereafter. We considered 70 articles out of the articles considered in the *'elsev_context_less'*. The first author has manually gone through each article to find implicit sentiment. He annotated each sentiment context with three labels viz. positive, negative and neutral. The dataset contains a total of 2164 citation sentences comprising positive, negative, and neutral sentences as 712, 316, and 1136 respectively. We observed that considering a bigger citation window introduces new negative sentiment bearing samples which was not the case for the previous dataset. Considering an example presented below, as per the *'elsev_context_less'* sentiment toward the citation *'23'* is positive but sentiment of the context-based sample toward the cited work is negative.

> *"Throughput per link to calculate the throughput under HT conditions, we take inspiration from the work of Cano etÃ, al. Ã, [23], where the HT problem in a WSN is undertaken. Unlike the work of Cano etÃ, al., we include diverse design premises of the IEEE 802.15.5 standard in the throughput characterization, which provide a more realistic study of the WMSN requirements."*

### 4.1.4 Athar dataset with context

To develop context full dataset, we considered 70 ACL Anthology articles out of 194 articles (Joseph & Radev, 2007), (Athar & Teufel, 2012). The first author has gone through each of the article manually to determine the context of each citation. Compared to 6569 citations sentences of *athar_context_less* dataset, context based dataset contains 1874 citation windows comprising 571 positive, 219 negative, and 1084 neutral samples. This dataset will be referred to as *athar_context_full* hereafter. Unlike *athar_context_less*, considering the context in scientific articles improved the quality of annotation as context helps in deciding the sentiment more precisely. We could not capture sentiments in single citation corpora in many cases. But, sentiments can be better captured using context based samples. It can be observed in the

---

[3] https://docs.python.org/3/library/xml.html.

following example:

> "Liang et al. (2009) simultaneously developed a method for learning with and actively selecting measurements or target expectations with associated noise. The measurement selection method proposed by Liang et al. (2009) is based on Bayesian experimental design and is similar to the expected information gain method described above. Consequently, this method is likely to be intractable for real applications."

In the above example, if single citation sentence is considered then sentiment becomes neutral, whereas the overall sentiment toward the cited work is positive.

We performed annotation keeping the importance of the context in mind. We considered the guidelines given in (Athar, 2014a) and (Ravi et al., 2018) to annotate citation sentences for our dataset. The process of annotation of implicit sentiment in the citation text is performed manually. The citation sentences which criticize the results, approach or performance of the cited work, is marked as negative citation. The citation that confirms, supports or uses the existing cited work is considered as positive citation. The citation that is neither negative nor affirmative is considered as neutral citation.

Let us consider some examples of positive implicit citation context:

> "This limitation is not strictly related to our approach, but to the optimization algorithm. However, as shown in [22], it is very straightforward to modify DAGAME in order to include the ability to take into account the usage of distinct resources. In fact, this extension of DAGAME is part of our future work."

In the stated example, irrespective of a limitation, when the cited work shows any modification for further use, the citation is labelled as positive.

> "The use case size point's method was evaluated in Braz and Vergilio (2006). The authors emphasized the internal

> *structure of the use case scenario in their method, where the primary actors take on roles and are classified based on an adjustment factor. This approach can lead to better evaluations of actors and use cases."*

The last sentence in the above context represents positive sentiment toward the cited work. If we consider a single sentence, the cited work may fall in the bucket of neutral sentiment label.

Let us consider examples of negative implicit sentiment in the citation context:

> *"Alpern etÃ, al. propose the Memory Hierarchy (MH) Framework [26] that reflects important practical considerations that are hidden by the RAM and HMM models: data are moved in fixed size blocks simultaneously at different levels in the hierarchy, and the memory capacity, as well as bus bandwidth, are limited at each level. But there are too many parameters in this model that can obscure algorithm analysis."*

Here, the author cited article (Sula & Miller, 2014) in the paper and expressed polite criticism toward it. In the process of annotation, we labelled such context with negative sentiment.

Negative annotations are not limited to such cases whereas, when the author tries to improve some part of the previous research then it shows a negative sentiment for the cited work and it is labelled in the same category.

> *"To overcome these drawbacks, Herda et al. [17] introduced a real-time method using an anatomical human model to predict the position of the markers. It is unfortunately very difficult and time-consuming to set up such a model."*

In the above citation context, sentiment toward cited work starts with a positive attitude but the overall sentiment considering the window of two sentences makes the sentiment negative. In addition, when the author suggests any improvement or shows that extra effort is required toward the cited work then after examining the context the sentiment is labelled as negative.

### 4.1.5 Augmented datasets

Augmented datasets were created to increase the sample sized. To create augmented datasets without context, we combined *elsev_context_less* and *athar_context_less* together. This dataset will be referred to as *augmented_context_less* hereafter. The dataset contains 15492 samples out of which 1969, 425 and 13101 are positive, negative and neutral respectively.

Augmented datasets with context were formed by combining *elsev_context_full* and *athar_context_full* together. This dataset will be referred to as *augmented_context_full* hereafter. With this dataset, we experimented with 4038 samples out of which 1283 are positive, 532 are negative and 2220 are neutral citation sentences under 10-fold cross-validation framework.

### 4.1.6 Derived datasets

In order to test the effect of context in CSA, we derived three context-less datasets from existing context-full datasets. Each of the derived context-less datasets contains single citation sentence based sample, which is taken from multiple sentences based citation sample. Hence, the number of samples of the derived datasets was same as the number of samples of context-full dataset. Similarly, the polarity of each of the sample of derived datasets was same as to context-full one.

## 4.2 Feature selection for traditional machine learning

To obtain features from each dataset, we employed StringToWordVector filter of Weka 3.8.0 to obtain DTM and considered TF-IDF weights. We employed $\chi^2$ and information gain feature subset selection to select the best 500, 250 and 100 features from unigram *(uni)*, bi-gram *(bi)*, tri-gram *(tri)*, *uni_bi* and *uni_bi_tri*. The attributes selected using chi-square will be referred to as *chi_500*, *chi_250* and *chi_100* hereafter. Similarly, attributes selected using information gain are named as *info_500*, *info_250* and *info_100*. The DTMs generated from feature subset selection methods were evaluated in WEKA experimenter. We considered grid search method to tune parameter for each classifier as explained in the next section.

## 4.3 Hyper parameter tuning details

The parameter tuning is a technique of regulating the elements which control the behaviour of the model. The default/strict parameters on every dataset do not promise the best prediction uniformly in all the cases. We employed a grid search approach to tune the parameter of the support vector machine for all the six datasets. At the first level, we tuned gamma (g) and cost (c) parameters of radial basis function kernel of support vector machine *(SVM-RBF)* (Hsu, Chang, Lin, & others, 2003) and linear kernel *(SVML)* as given in Table 3. The best parameter values obtained at the first level were further tuned. If the best parameter value for g or c was *x*, then we considered $2^{(x-1)}$ to $2^{(x+1)}$ with the step size 2 to determine the best parameter value at the next level.

**Table 3: Range of parameter of SVM model used in grid search**

| Model Name | Parameter name | Default Value | Range | Step size |
|---|---|---|---|---|
| SVML | Cost (c) | 1 | $2^c$ where c = -3 to 15 | 2 |
| SVM-RBF | Cost (c) | 1 | $2^c$ where c = -3 to 15 | 2 |
|  | Gamma (g) | 0 | $2^g$ where g = -15 to 3 | 2 |

For wvCNN, we experimented with a mini-batch size of 10, 20, 30, and 50. We experimented with the number of feature maps *(fltr)* as 40, 100, and 128. We tried dropout rate of 0.2 at embedding layer *(DRe)* and 0.3 and 0.5 at pooling layer *(DRp)*. The embedding dimension (em_dim) was fixed at 300. We experimented with the filter window (fltr_win) size of 3 and 4. We experimented with the sequence length *(seq_len)* of 100, 150, 200 and 300. The sequence length refers to the maximum number of words considered as input at the embedding layer. The architecture is trained for 10, 25, and 50 epochs. Experiments using wvCNN are performed on Nvidia 1060.

For Oh-CNN and Oh-biLSTMp, we experimented with the different mini-batch sizes of 10, 20, 30, 40, 50, and 100. The number of nodes at the pooling layer was tried for 200, 250, 500, and 1000. The dropout rate of 0.5 was tried at the top layer *(DRt)* and 0.3 and 0.5 at *DRp*. The region size was tried for 3 and 5. The number of epochs experimented with 50 and 100. The step size decay was tried for 0.1, 0.2, and 0.3. Finally, the L2 regularization *(L2)* value was tried for 0, 1e-2, 1e-4, and 1e-6. The rest of the parameters and settings were same as that of (Johnson & Zhang, 2014), (Johnson & Zhang, 2016). Experiments using Oh-CNN and Oh-biLSTMp were performed on Nvidia 1060.

## 4.4 Evaluation metrics

To compare performance of different models, we reported average macro-F1 score using 10-fold cross-validation. The micro-F1 and macro-F1 are commonly used performance measures for multi-classification task. The micro-F1 (weighted average) becomes problematic when the dataset suffers from imbalance issue. The weighted average would be biased towards the class which outnumber the instances of other classes. In our case, neutral citation sentences outnumber the instances of positive and negative instances. The bias can be adjusted using the macro-F1 score.

The precision denotes the ratio of correctly predicted positive observations *(TP)* to the total predicted positive observations *(TP + FP)*, high precision typically corresponds to low false positive predictions. Its formulation is presented in equation (1)

$$Precision = \frac{TP}{TP+FP} \tag{1}$$

Here, *TP, FP, TN,* and *FN* are abbreviated form of true positive, false positive, true negative, and false negative respectively.

The recall is defined as the ratio of correctly labelled positive observations to all the positive observations in the actual class. It measures the ability of the trained model in identifying positive observations from all the samples that should have been labelled as positive. Mathematically, it is presented in equation (2)

$$Recall = \frac{TP}{TP+FN} \tag{2}$$

It is the harmonic mean between precision and recall. The range is 0 to 1. A larger value indicates better predictive accuracy. Its formulation is given in equation (3)

$$Macro - F1 = 2 \times \frac{precision \times recall}{precision+recall} \tag{3}$$

In the macro-F1 score, the relative contribution of precision and recall is the same.

## 5 Results and discussion

We followed a grid search approach to tune the hyper-parameters of different classifiers. We experimented with two kinds of dataset viz. context-full and context-less. As we created our own dataset in the case of context-full, we could not compare our results with any other literature. In the case of context-less, we performed all the experiments without balancing, whereas (Ravi et al., 2018) performed all the experiments after balancing the dataset. Hence, we compared our results with that of (Ravi et al., 2018) for the sake of brevity. To do this, we considered the same number of test samples of each fold of 10-fold cross validation (FCV) as that of (Ravi et al., 2018).

Table 4 presents comparison of results obtained by (Ravi et al., 2018) on balanced dataset with our results obtained on non-balanced dataset. Here, SVM (RBF) yielded the best macro-F1 of 63.76% without balancing the *elsev_context_less* dataset, whereas (Ravi et al., 2018) reported the best macro-F1 of 54.49% obtained using Oh-biLSTMp after balancing the dataset. We achieved 10.27% better macro-F1 than that of (Ravi et al., 2018).

Among other deep learning methods, wvCNN_non_static yielded the highest macro-F1 of 55.76%. Traditional machine learning approach outperformed deep learning methods by macro-F1 of 8%. We performed t-test between the best results obtained using SVM (RBF) vs. the rest of classifiers at 1% significance level and 18 degrees of freedom. The t-test indicated that the results obtained using SVM (RBF) is statistically significantly better than the rest of classifiers. The obtained results help to infer that the number of samples are not sufficient for deep learning to yield better results than traditional deep learning methods.

Table 4: Comparison of results with (Ravi et al., 2018) for context-less dataset

| Techniques | Macro-F1 for Elsevier dataset without context | | | Macro-F1 for ACL dataset without context | | |
|---|---|---|---|---|---|---|
| | (Ravi et al., 2018) on balanced dataset | Ours results on non-balanced *elsev_context_less* | t-statistic between SVM(RBF) vs. all on our results | (Ravi et al., 2018) on balanced dataset | Ours results on non-balanced *athar_context_less* | t-statistic between Oh_biLSTMp vs. all on our results |
| SVM (RBF) | 37.12% | **63.76%** | - | 62.99% | 62.81% | 5.02 |
| wvCNN_random | 46.4% | 55% | 6.71 | 50.7% | 51.74% | 12.39 |
| wvCNN_non-static | 44.44% | 55.76% | 4.21 | 49.37% | 52% | 10.56 |
| Oh_CNN | 53.87% | 43.03% | 15.77 | 51.9% | **66.54%** | 2.61 |
| Oh_biLSTMp | 54.49% | 46.8% | 17.21 | 58.16% | **70.14%** | - |

For *athar_context_less* dataset, our approach outperformed each model used in (Ravi et al., 2018). (Ravi et al., 2018)obtained the best results for balanced *athar_context_less* dataset using SVM and reported macro-F1 of 62.99%. In our experiments, Oh-biLSTMp yielded the best macro-F1 of 70.14%, which is 7.15% better than that of (Ravi et al., 2018). For our results, we performed two-tailed t-test between the results obtained using Oh-biLSTMp versus the rest of classifiers at 1% level of significance and 18 degrees of freedom. Based on t-test, we observed that Oh_CNN yielded macro-F1 of 66.54%, which is statistically significantly better than the rest of classifiers. Region embedding based both models yielded better results than word vector based CNN and SVM. The possible reason behind obtaining better results on *athar_context_lessl* dataset using region-embedding based models would be the characteristics of the dataset. All samples of *athar_context_less* dataset are taken from the same subject i.e. computational linguistics. In the case of *elsev_context_less* dataset, samples are taken from different subject areas as mentioned in Section 4.1.1. This indicates that the region-embedding based models yield better results for the dataset containing phrases of similar semantic abundantly.

Table 5 presents the results obtained for *elsev_context_full* dataset. Here, SVM (RBF) yielded the best macro-F1 of 67.74% using $\chi^2$ based 500 features. Among deep learning methods, wvCNN_random yielded the highest macro-F1 of 66.3%. Traditional machine learning approach is able to outperform deep learning methods by improved macro-F1 of 1.44%. We performed two-tailed t-test between the results obtained using SVM (RBF) versus all at 1% level of significance and 18 degrees of freedom. The t-test indicates that SVM (RBF) is statistically significantly same as the rest of the classifier. Based on these results, we can infer that *'tri'* features are helpful for SVM as well as deep learning methods.

Results obtained from *athar_context_full* dataset are reported in Table 6. SVM (RBF) yielded the best macro-F1 of 77.33% using $\chi^2$ based 500 *'bi'* features. Among deep learning methods, Oh-biLSTMp yielded the highest macro-F1 of 66.9%. Traditional machine learning approach is able to outperform deep learning methods by macro-F1 of 10.43%. We performed two-tailed t-test between the results obtained using SVM (RBF) versus all at 1% level of significance and 18 degrees of freedom. Based on the t-test, we observed that SVM (RBF) statistically significantly outperformed rest of the classifier.

The results obtained by the framework on context-full dataset are quite promising. It is observed that context-enhanced citation sentences played an important role in model training. Further, the issues mentioned in (Ravi et al., 2018) viz. granularity, the vagueness of balance, indirect narration, and multiple citations are resolved in the development of context-full dataset manually. The problem of multiple citations within the same sentence is tackled because we have better clarity of opinion orientation in context rich citation sentences.

Table 6 indicates that traditional machine learning approach outperformed other models, whereas the results reported by our approach are better than (Ravi et al., 2018). For context-full datasets, we have small amount of samples owing to expensive manual annotation of samples.

Even though, our developed context-full dataset is small, the results obtained seems promising for CSA. As (Ravi et al., 2018) reported, increasing the number of samples for deep learning would improve the CSA results. To test this hypothesis, we clubbed the context-less datasets taken from Elsevier and ACL together. Similarly, we augmented context-full datasets together too.

The results obtained on *augmented_context_less* dataset is reported in Table 7. SVM (RBF) yielded the best macro-F1 of 63.31%. It is observed that increasing the sample size by combining context less dataset did not perform well. The reason behind the low score is the quality of citation sentences in the dataset. It is noticeable that when the single citation sentence or fixed sentence window is considered to train a model, the learning is not effective. Table 8 presents results obtained on *augmented_context_full* dataset. Here, Oh-biLSTMp yielded macro-F1 of 69%. Based on t-test, we observed that Oh_CNN which yielded macro-F1 of 68% and SVM (RBF) with macro-F1 of 64.85% are statistically significantly same as Oh-biLSTMp. We can observe that traditional as well as deep learning models yield better performance using higher number of context enhanced samples. With merely 26% samples as compared to *augmented_context_less* dataset, we obtained better results on *augmented_context_full* as reported in Table 8.

Table 5: Macro-F1 obtained for elsev_context_full dataset

| Techniques | Parameter | Macro-F1 (%) | t-statistic |
|---|---|---|---|
| SVM (RBF) | tri, chi_500, c = 32768, g = 0.625 | **67.74** | - |
| wvCNN_random | Batch size = 50, fltr = 100, DRe = 0.2, DRp = 0.5, em_dim = 300, fltr_win = (3,4), seq_len = 300, Epoch = 10 | **66.3** | **0.62** |
| wvCNN_non-static | Batch size = 50, fltr = 100, DRe = 0.2, DRp = 0.5, em_dim = 300, fltr_win = (3,4), seq_len = 300, Epoch = 10 | **61.8** | **0.85** |
| Oh-CNN | Batch size = 100, Nodes = 1000, Drt = 0.5, Epoch = 100, Decay = 0.1, L2 = 1e-4 | **65.95** | **1.14** |
| Oh-biLSTMp | Batch size = 50, Nodes = 500, Drt = 0.5, Epoch = 50, Decay = 0.3, L2 = 0 | **64.7** | **1.79** |

Table 6: Macro-F1 obtained for athar_context_full dataset

| Techniques | Parameter | Macro-F1 (%) | t-statistic |
|---|---|---|---|
| SVM (RBF) | bi, chi_500, c = 32, g = 0.625 | **77.33** | - |
| wvCNN_random | Batch size = 50, fltr = 100, DRe = 0.2, DRp = 0.5, em_dim = 300, fltr_win = (3,4), seq_len = 300, Epoch = 10 | 54.03 | 8.48 |
| wvCNN_non-static | Batch size = 50, fltr = 100, DRe | 50.76 | 10.55 |
| | = 0.2, DRp = 0.5, em_dim = 300, fltr_win = (3,4), seq_len = 300, Epoch = 10 | | |
| Oh-CNN | Batch size = 100, Nodes = 1000, Drt = 0.5, Epoch = 100, Decay = 0.1, L2 = 1e-4 | 65.2 | 6.18 |
| Oh-biLSTMp | Batch size = 50, Nodes = 500, Drt = 0.5, Epoch = 50, Decay = 0.3, L2 = 0 | 66.9 | 5.36 |

**Table 7: Macro-F1 obtained for augmented_context_less dataset**

| Techniques | Parameter | Macro-F1 (%) | t-statistic |
|---|---|---|---|
| SVM (RBF) | uni, chi_500, c = 8192, g = 6.11e-4 | **63.31** | - |
| wvCNN_random | Batch size = 50, fltr = 100, DRe= 0.2, DRp = 0.5, em_dim = 300, fultr_win = (3,4), seq_win = 300, Epoch = 10 | 56.85 | 5.45 |
| wvCNN_non-static | Batch size = 50, fltr = 100, DRe = 0.2, DRp = 0.5, em_dim = 300, fltr_win = (3,4), seq_len = 300, Epoch = 10 | 58.05 | 4.66 |
| Oh-CNN | Batch size = 100, Nodes = 1000, Drt = 0.5, Epoch = 100, Decay = 0.1, L2 = 1e-4 | 58.1 | 3.37 |
| Oh-biLSTMp | Batch size = 50, Nodes = 500, Drt = 0.5, Epoch = 50, Decay = 0.3, L2 = 0 | 59.2 | 3.66 |

**Table 8: Macro-F1 obtained for augmented_context_full dataset**

| Techniques | Parameter | Macro-F1 (%) | t-statistic |
|---|---|---|---|
| SVM (RBF) | uni_bi_tri, chi_250, c = 8192, g = 6.11e-4 | **64.85** | 2.91 |
| wvCNN_random | Batch size = 50, fltr = 100, DRe = 0.2, DRp = 0.5, em_dim = 300, fltr_win = (3,4), seq_len = 300, Epoch = 10 | 55.78 | 9.07 |
| wvCNN_non-static | Batch size = 50, fltr = 100, DRe = 0.2, DRp = 0.5, em_dim = 300, fltr_win = (3,4), seq_len = 300, Epoch = 10 | 52.99 | 11.79 |
| Oh-CNN | Batch size = 100, Nodes = 1000, Drt = 0.5, Epoch = 100, Decay = 0.1, L2 = 1e-4 | **68** | 0.53 |
| Oh-biLSTMp | Batch size = 50, Nodes = 500, Drt = 0.5, Epoch = 50, Decay = 0.3, L2 = 0 | **69** | - |

**Table 9: Macro-F1 obtained for elsev_context_full_derived dataset**

| Techniques | Parameter | Macro-F1(%) | t-statistic |
|---|---|---|---|
| SVM (RBF) | uni_bi_tri, chi_500, c = 2, g = 0.0039 | **62.16** | - |
| wvCNN_random | Batch size = 50, fltr = 100, DRe= 0.2, DRp = 0.5, em_dim = 300, fltr_win = (3,4), seq_win = 300, Epoch = 10 | 41.68 | 9.85 |
| wvCNN_non-static | Batch size = 50, fltr = 100, DRe = 0.2, DRp = 0.5, em_dim = 300, fltr_win = (3,4), seq_len = 300, Epoch = 10 | 33.33 | 18.29 |
| Oh-CNN | Batch size = 100, Nodes = 1000, Drt = 0.5, Epoch = 100, Decay = 0.1, L2 = 1e-4 | 39.33 | 15.98 |
| Oh-biLSTMp | Batch size = 50, Nodes = 500, Drt = 0.5, Epoch = 50, Decay = 0.3, L2 = 0 | 45.92 | 10.62 |

**Table 10: Macro-F1 obtained for athar_context_full_derived dataset**

| Techniques | Parameter | Macro-F1 (%) | t-statistic |
|---|---|---|---|
| SVM (RBF) | uni_bi_tri, chi_500, c = 2, g = 0.0039 | **62.97** | - |
| wvCNN_random | Batch size = 50, fltr = 100, DRe= 0.2, DRp = 0.5, em_dim = 300, fltr_win = (3,4), seq_win = 300, Epoch = 10 | 46.58 | 6.05 |
| wvCNN_non-static | Batch size = 50, fltr = 100, DRe = 0.2, DRp = 0.5, em_dim = 300, fltr_win = (3,4), seq_len = 300, Epoch = 10 | 39.45 | 8.13 |
| Oh-CNN | Batch size = 100, Nodes = 1000, Drt = 0.5, Epoch = 100, Decay = 0.1, L2 = le-4 | 37.05 | 12.87 |
| Oh-biLSTMp | Batch size = 50, Nodes = 500, Drt = 0.5, Epoch = 50, Decay = 0.3, L2 = 0 | 49.9 | 5.95 |

**Table 11: Macro-F1 obtained for augmented_context_full_derived dataset**

| Techniques | Parameter | Macro-F1 (%) | t-statistic |
|---|---|---|---|
| SVM (RBF) | uni_bi_tri, chi_500, c = 2, g = 0.0039 | **51.59** | **1.07** |
| wvCNN_random | Batch size = 50, fltr = 100, DRe= 0.2, DRp = 0.5, em_dim = 300, fltr_win = (3,4), seq_win = 300, Epoch = 10 | 49.45 | 3.34 |
| wvCNN_non-static | Batch size = 50, fltr = 100, DRe = 0.2, DRp = 0.5, em_dim = 300, fltr_win = (3,4), seq_len = 300, Epoch = 10 | 41.92 | 10.18 |
| Oh-CNN | Batch size = 100, Nodes = 1000, Drt = 0.5, Epoch = 100, Decay = 0.1, L2 = le-4 | 49.77 | 4.16 |
| Oh-biLSTMp | Batch size = 50, Nodes = 500, Drt = 0.5, Epoch = 50, Decay = 0.3, L2 = 0 | **52.98** | - |

**Table 12: Comparison between context-full and their respective derived datasets**

| Source | Dataset | Model | Macro-F1 (%) | t-statistic |
|---|---|---|---|---|
| Elsevier | elsev_context_full | SVM (RBF) | **67.66** | - |
|  | elsev_context_full_derived | SVM (RBF) | 62.16 | 3.43 |
| ACL | athar_context_full | SVM (RBF) | **77.32** | - |
|  | athar_context_full_derived | SVM (RBF) | 62.97 | 5.4 |
| Augmented | augmented_context_full | Oh_LSTMp | **69** | - |
|  | augmented_context_full_derived | Oh_LSTMp | 52.98 | 11.77 |

The results of three derived datasets are presented in Table 9, 10, and 11. For *elsev_context_full_derived* dataset, SVM (RBF) yielded the best macro-F1 of 62.16%, which is statistically significantly better than the rest of the classifiers. For *athar_context_full_derived* dataset, SVM (RBF) yielded the best macro-F1 of 62.97%, which is statistically significantly better than the rest of the classifiers. For *augmented_context_fu- ll_derived* dataset, Oh-biLSTMp yielded the best macro-F1 of 52.98%, which is statistically significantly same as SVM (RBF). Comparison between context-full and its derived datasets are presented in Table 12. We can observe that we are able to achieve better results using context-full datasets in all the three cases. Hence, we can conclude that context plays significant role in CSA.

# 6 Conclusions

The work in this paper was carried out for qualitative analysis of citation sentences to derive sentiment of author regarding the cited work. The new context enhanced dataset for CSA and feature selection methods are found to be useful in dealing with data imbalance issue. Our approach with very less samples in context enhanced dataset yielded promising results. We observed that using region embedding LSTM and CNN methods yielded better results on the datasets generated from scientific articles collected from similar background. Furthermore, deep learning methods are statistically significantly same as traditional machine learning methods for citation sentiment anlaysis tasks. We also observed that deep learning methods are suitable for higher number of samples, whereas traditional machine learning methods are suitable for smaller number of samples. We also observed that we can achieve higher performance on context-full datasets compared to context-less datasets.

In future, we would create a bigger annotated corpora of context-full citations. Further, citation sentence context extraction and polarity annotation are very challenging as well as expensive tasks. We need to develop techniques to perform these tasks automatically. Novel deep learning model needs to be developed to perform citation sentiment classification, which should be less compute and time intensive.

**Disclaimer:** This paper represents the opinions of the authors, and is the product of academic research. It is not meant to represent the views, thoughts, and opinions of author's employer, organization, committee or other group or individual. Any errors are the fault of the authors.

# References

Athar, A. (2011). Sentiment analysis of citations using sentence structure-based features. In *Proceedings of the ACL 2011 student session* (pp. 81–87).

Athar, A. (2014a). *Sentiment analysis of scientific citations, Ph.D. Thesis*. University of Cambridge.

Athar, A. (2014b). Sentiment analysis of scientific citations. Retrieved from http://www.cl.cam.ac.uk/

Athar, A., & Teufel, S. (2012). Context-Enhanced Citation Sentiment Detection. In *Proceedings of the 2012 Conference of the North American Chapter of the Association for Computational Linguistics: Human Language Technologies* (pp. 597–601). USA: Association for Computational Linguistics.

Bu, Y., Waltman, L., & Huang, Y. (2019). A multidimensional perspective on the citation impact of scientific publications. *CoRR*, *abs/1901.0*. Retrieved from http://arxiv.org/abs/1901.09663

De Marneffe, M.-C., & Manning, C. D. (2008). *Stanford typed dependencies manual*.

Hsu, C.-W., Chang, C.-C., Lin, C.-J., & others. (2003). A practical guide to support vector classification.

Ikram, M. T., & Afzal, M. T. (2019). Aspect based citation sentiment analysis using linguistic patterns for better comprehension of scientific knowledge. *Scientometrics*, *119*(1), 73–95.

Johnson, R., & Zhang, T. (2014). Effective use of word order for text categorization with convolutional neural networks. *ArXiv Preprint ArXiv:1412.1058*.

Johnson, R., & Zhang, T. (2016). Supervised and semi-supervised text categorization using LSTM for region embeddings. In *International Conference on Machine Learning* (pp. 526–534).

Joseph, M. T., & Radev, D. R. (2007). *Citation analysis, centrality, and the ACL anthology*. *Technical Report CSE-TR-535-07*.

Kalchbrenner, N., Grefenstette, E., & Blunsom, P. (2014). A convolutional neural network for modelling sentences. *ArXiv Preprint ArXiv:1404.2188*.

Karyotis, C., Doctor, F., Iqbal, R., James, A., & Chang, V. (2018). A fuzzy computational model of emotion for cloud based sentiment analysis. *Information Sciences*, *433–434*, 448–463. https://doi.org/10.1016/j.ins.2017.02.004

Kim, Y. (2014). Convolutional Neural Networks for Sentence Classification. In *Proceedings of the 2014 Conference on Empirical Methods in Natural Language Processing ({EMNLP})* (pp. 1746–1751). Doha, Qatar: Association for Computational Linguistics. https://doi.org/10.3115/v1/D14-1181

Lauscher, A., Glavaš, G., Ponzetto, S. P., & Eckert, K. (2017). Investigating convolutional networks and domain-specific embeddings for semantic classification of citations. In *Proceedings of the 6th International Workshop on Mining Scientific Publications* (pp. 24–28).

Loper, E., & Bird, S. (2002). NLTK: the natural language toolkit. *ArXiv Preprint Cs/0205028*.

Maks, I., & Vossen, P. (2012). A lexicon model for deep sentiment analysis and opinion mining applications. *Decision Support Systems*, *53*(4), 680–688.

Mikolov, T., & Dean., J. (2013). Distributed representations of words and phrases and their compositionality. In *Advances in neural information processing systems* (pp. 1–9).

Munkhdalai, T., Lalor, J., & Yu, H. (2016). Citation analysis with neural attention models. In *Proceedings of the Seventh International Workshop on Health Text Mining and Information Analysis* (pp. 69–77).

Parthasarathy, G., & Tomar, D. C. (2014). Sentiment analyzer: Analysis of journal citations from citation databases. In *Confluence The Next Generation Information Technology Summit (Confluence), 2014 5th International Conference-* (pp. 923–928).

Pennington, J., Socher, R., & Manning, C. (2014). Glove: Global vectors for word representation. In *Proceedings of the 2014 Conference on Empirical Methods in Natural Language Processing (EMNLP)* (pp. 1532–1543).

Radev, D. R., Muthukrishnan, P., Qazvinian, V., & Abu-Jbara, A. (2013). The ACL anthology

network corpus. *Language Resources and Evaluation*, *47*(4), 919–944.

Ravi, K., Setlur, S., Ravi, V., & Govindaraju, V. (2018). Article citation sentiment analysis using deep learning. In *2018 IEEE 17th International Conference on Cognitive Informatics & Cognitive Computing (ICCI\* CC)* (pp. 78–85).

Sula, C. A., & Miller, M. (2014). Citations, contexts, and humanistic discourse: Toward automatic extraction and classification. *Literary and Linguistic Computing*, *29*(3), 452–464.

Toutanova, K., Klein, D., Manning, C. D., & Singer, Y. (2003). Feature-rich part-of-speech tagging with a cyclic dependency network. In *Proceedings of the 2003 conference of the North American chapter of the association for computational linguistics on human language technology-volume 1* (pp. 173–180).

Wang, M., Leng, D., Ren, J., Zeng, Y., & Chen, G. (2019). Sentiment classification based on linguistic patterns in citation context. *CURRENT SCIENCE*, *117*(4), 606.

White, H. D. (2004). Citation analysis and discourse analysis revisited. *Applied Linguistics*, *25*(1), 89–116.

Zaremba, W., & Sutskever, I. (2014). Learning to execute. *ArXiv Preprint ArXiv:1410.4615*.

Zhao, D., & Strotmann, A. (2020). Deep and narrow impact: introducing location filtered citation counting. *Scientometrics*, *122*(1), 503–517.